\pdfoutput=1
\documentclass[letterpaper, 10 pt, conference]{ieeeconf}  

\IEEEoverridecommandlockouts                              

\usepackage{times}

\usepackage{xcolor}
\usepackage{soul}
\usepackage[utf8]{inputenc}
\usepackage[small]{caption}

\usepackage{graphicx}
\usepackage[caption=false, font=footnotesize]{subfig}
\usepackage[export]{adjustbox}

\usepackage{multicol}
\usepackage[bookmarks=true]{hyperref}

\usepackage{pgfplots}
\pgfplotsset{width=7.5cm, compat=1.9}
\usepackage{amsmath}

\usepackage{algorithm}
\usepackage[noend]{algpseudocode}

\usepackage{color}
\newcommand{\adnote}[1]{\textcolor{blue}{Anca: #1}}

\title{\LARGE \bf
Agent Prioritization for Autonomous Navigation
}

\author{Khaled S. Refaat$^{*1}$, Kai Ding$^{*1}$, Natalia 
Ponomareva$^{2}$, and St\'ephane Ross$^{1}$ 
\thanks{*indicates equal contribution.}
\thanks{$^{1}$Khaled S. Refaat, Kai Ding and St\'ephane Ross are with Waymo, Mountain View, CA 94043, USA.
        {\tt\small \{krefaat, dingkai, stephaneross\}@waymo.com.}}%
\thanks{$^{2}$Natalia Ponomareva is with Google Research, Mountain View, CA 94043, USA.
        {\tt\small nponomareva@google.com.}}%
}

\begin{document}

\maketitle
\thispagestyle{empty}
\pagestyle{empty}

\begin{abstract}

In autonomous navigation, a planning system reasons about other agents to plan a safe and plausible trajectory.
Before planning starts, agents are typically processed with computationally intensive models for recognition, tracking, motion estimation and prediction.
 With limited computational resources and a large number of agents to process in real time, it becomes important to efficiently rank agents according to their impact on the decision making process.
 This allows spending more time processing the most important agents. We propose a system to rank agents around an autonomous vehicle (AV) in real time.
We automatically generate a ranking data set by running the planner in simulation on real-world logged data, where we can afford to run more accurate and expensive models on all the agents. The causes of various planner actions are logged and used for assigning ground truth importance scores. The generated data set can be used to learn ranking models. In particular, we show the utility of combining learned features, via a convolutional neural network, with engineered features designed to capture domain knowledge.
We show the benefits of various design choices experimentally. When tested on real AVs, our system demonstrates the capability of understanding complex driving situations.

\end{abstract}

\section{INTRODUCTION}

An autonomous vehicle (AV) uses a variety of sensors such as radars, lidars and cameras to detect agents (e.g. other vehicles, cyclists and pedestrians) in the scene~\cite{Urmson:2008:SCU:1399090.1399250, DBLP:journals/corr/BojarskiTDFFGJM16, Ogale-RSS-19}. To be able to plan a safe and comfortable trajectory for the AV, its onboard software typically needs to process the detected agents using computationally intensive models. For example, the onboard software may need to predict what other agents are going to do in the next few seconds~\cite{Alahi_2016_CVPR}.
It may also utilize expensive models for tracking, recognition (e.g. poses, gestures, blinkers) and motion estimation applied to individual agents; e.g.~\cite{hochreiter1997long,rudenkoICRAPalmieri2018,journals/pami/WuLY15,Held-2017-103049,8206288}. The output of this processing is then provided as input to a planning module, which outputs a trajectory for the AV to follow.

With limited computational resources and a large number of agents in the scene to process, the onboard software can not use computationally intensive models for all the agents. A natural approach to tackle this problem is to prioritize among the agents and spend the majority of time reasoning about the most important agents, which are likely to affect the planner decisions. This is a very challenging task since in order to rank agents, one may need to reason about potential interactions with them. Humans perform well on this task, quickly focusing on a small number of vehicles while driving and ignoring many others, for example vehicles driving far behind or heading away.

\begin{figure}[]
\center
\includegraphics[scale=.55,clip=true,angle=0]{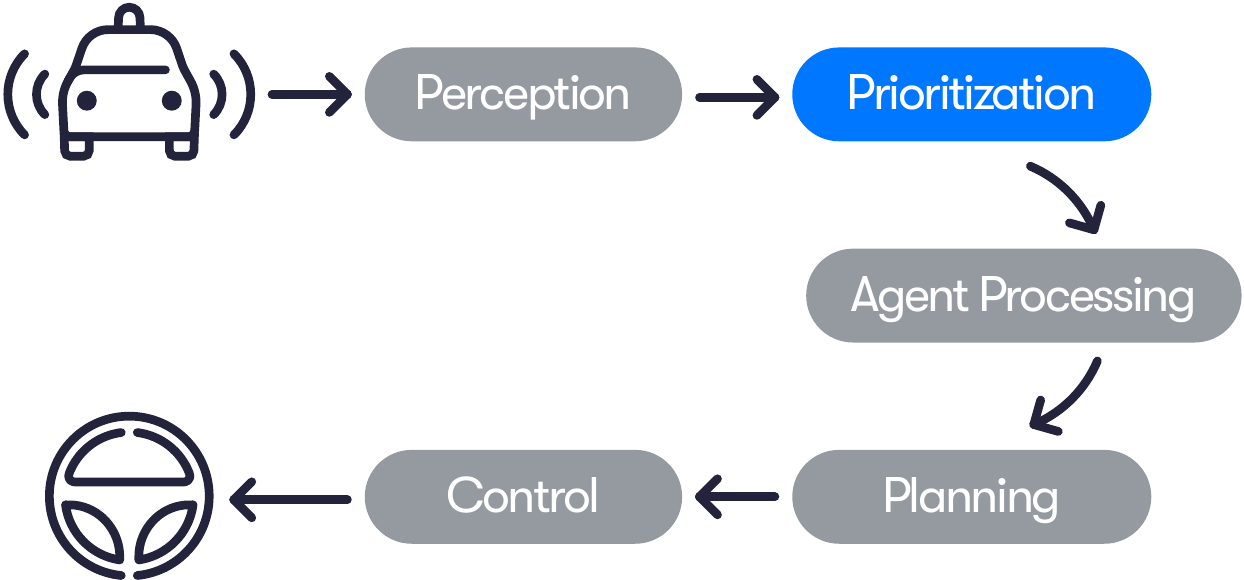}
\caption{A simplified block diagram of an autonomous navigation system highlighting the role of a prioritization module which ranks agents according to their influence on the planner's decision making. The most important agents are then processed using more computationally intensive models and algorithms than those used for other agents.}
\label{fig:block}
\end{figure}

Figure~\ref{fig:block} illustrates a block diagram of an autonomous navigation system showing the role of a prioritization module. The perception system processes sensor data to detect agents in the scene. The prioritization module ranks the detected agents according to their importance to the planner's decision making procedure. The most important agents are potentially processed using more computationally intensive models and algorithms to provide the planner with useful information. Other agents are still being processed using more efficient, and potentially less accurate, models.
The planner module outputs a driving trajectory for the AV, which is converted to control commands to be executed in real time.

While simple heuristics could be used to rank agents, for example, based on their arrival time to the AV's intended path, we found that learning a ranking model from data leads to more accurate results. Fortunately, as we will show, one could generate data automatically without any human labeling. However, one challenge is that we need to learn from data that captures complex interactions between the AV and its surrounding agents, which suggests training based on perceptual outputs over time.
Another challenge is that the ranking may change abruptly due to the appearance of new agents or the sudden reckless behavior of others. Thus, the ranking model has to be efficient enough to allow frequent re-ranking, for example before every planning iteration\footnote{Information from the previous cycle (e.g reactions and previous ranks) could be used to {\it{directly}} rank agents. However, this approach may lead to poor handling of sudden behavioral changes and newly detected agents.}.

One way to tackle this problem is to hand-design a set of engineered features that describe every agent behavior, and use a machine learning model to output an importance score which could be used for ranking. Engineered features are often hard to design and may result in a significant loss of information about the dynamic scene.
A more attractive approach is to use a deep neural network that consumes the raw data to output importance scores. However, deep networks that consume data over long temporal intervals may not be efficient enough to be frequently called onboard an AV in real time.

{\it{Contributions:}} Real-time agent prioritization onboard an autonomous vehicle, using a model learned to rank, is novel to the best of our knowledge.
We propose a system to rank agents around an AV in real time according to their importance to the decision making of the AV. Our system consists of multiple components:

\subsubsection{Data Generation} We automatically generate ground truth ranking data without human intervention. We run the planner in simulation on real-world logged data, and log the causes of different planner decisions such as braking and nudging.  These causes are then processed off-board to assign ground truth importance scores to agents, which are used to learn ranking models.

\subsubsection{Data Representation} We represent perceptual outputs through time (e.g. agent locations, speed and acceleration profiles) compactly to provide input to a deep convolutional neural network (CNN)~\cite{LeCun:1999:ORG:646469.691875}. The main idea is to represent the time information as a separate channel which, when combined with other input channels, provides information about the scene as it changes through time; see Figures~\ref{fig:corr} and~\ref{fig:images}.

\subsubsection{Hybrid Approach}
We design a model based on gradient boosted decision trees (GBDT), which uses manually designed features, in addition to features extracted from the output of a CNN. The GBDT optimizes a pairwise loss to predict an importance score for each agent. The CNN captures complex interactions between the AV and the surrounding agents that are otherwise hard to capture using manually designed features. We analyze the benefits of the hybrid approach and the pairwise loss experimentally.

\section{RELATED WORK}
\subsection{Agent Processing}
In autonomous vehicles, processing agents with expensive models is common. In agent trajectory prediction, Alahi {\it{et al}}~\cite{Alahi_2016_CVPR} showed how to predict future trajectories by having an LSTM~\cite{hochreiter1997long} for each person in a scene. Rudenko {\it{et al}}~\cite{rudenkoICRAPalmieri2018} presented a planning-based algorithm for long-term human motion prediction that takes local interactions into account. In scenarios involving only ten people, the average run time of the algorithm ranges from 0.5 to 1.2 seconds for prediction horizons ranging from 7.5 to 15 seconds. Wu {\it{et al}}~\cite{journals/pami/WuLY15} reviewed recent advances in object tracking and showed that most algorithms used to track individual agents have a speed under 100 frames per second. Held {\it{et al}}~\cite{Held-2017-103049} proposed a tracking method that uses a pre-trained feed-forward network with no online training required, and the tracker is able to run at 6.05 to 9.98 milliseconds per frame with GPU acceleration. On pose estimation, Chu {\it{et al}}~\cite{Chu2017MulticontextAF} proposed an end-to-end framework with a multi-context attention convolutional neural network and a conditional random field, for single-person human pose estimation from camera images. For action recognition, Zhao {\it{et al}}~\cite{8206288} presented a voting approach, combining 3D-CNNs with bidirectional gated RNNs to recognize actions from video sequences. The execution time of these methods typically grows linearly with the number of agents which calls for an efficient ranking system to prioritize among agents in real time~\footnote{Distributed computing can also be used but with limited resources onboard an AV, ranking becomes important.}. Agents that are less likely to change the AV behavior can be processed with more efficient, and potentially less accurate, models.

\subsection{Learning to Rank}
Learning to rank has played a crucial role in applications such as information retrieval while typically relying on human-labeled data:
for example, Cohen {\it{et al}}~\cite{NIPS1997_1431} considered the problem of learning how to order, given feedback in the form of preference judgments. They outlined a two-stage approach in which conventional means are used to learn a preference function, and new instances are
then ordered to maximize agreements with the learned preference function. Burges {\it{et al}}~\cite{Burges05} introduced RankNet which uses gradient descent to optimize a differentiable objective function that penalizes the model for outputting query scores disagreeing with the ground truth ranks. To tackle the problem of the discrepancy between the target and the optimization costs of the ranking problem, Burges {\it{et al}}~\cite{NIPS2006_2971}  proposed defining a virtual gradient on each item in the list after sorting. On the other hand, Valizadegan {\it{et al}}~\cite{NIPS2009_3758} solved this issue by optimizing directly for the  expectation of the Normalized Discounted Cumulative Gain (NDCG), which is a popular measure in information retrieval~\cite{Jarvelin:2000:IEM:345508.345545}. Other systems such as~\cite{NIPS2007_3270} use multiple classification as a proxy for optimizing for NDCG, relying on the observation that perfect classifications result in perfect NDCG scores. 

 Jamieson and Nowak~\cite{NIPS2011_4427} showed how to rank a set of objects using pairwise comparisons assuming the objects could be embedded in an Euclidean space.
 Yun {\it{et al}}~\cite{NIPS2014_5363} proposed RoBiRank which is a ranking algorithm motivated by observing a connection between evaluation metrics for learning to rank, and loss functions for robust classification which operate in the presence of outliers.

Human labels were previously used to learn off-board machine learning models that classify road users according to their importance, or generate a saliency map, from driving videos; e.g.~\cite{Ohn-Bar:2017:OED:3030484.3030516,8569438}.
To rank agents around an AV in real time, we utilize gradient boosted decision trees and convolutional neural networks.

\subsection{Gradient Boosted Decision Trees}
Gradient Boosted Decision Trees (GBDT) have gained a lot of interest in recent years. Initially introduced by Friedman~\cite{friedman2001greedy}, they have become frequent winners in Kaggle and KDD Cup competitions~\cite{chen2016xgboost}. Several open-sourced implementations were released, such as scikit-learn~\cite{pedregosa2011scikit}, R gbm~\cite{Ridgeway05generalizedboosted}, LightGBM~\cite{ke2017lightgbm}, XGBoost~\cite{chen2016xgboost} and TensorFlow Boosted Trees~\cite{ponomareva2017tf}. In our implementation, we use~\cite{ponomareva2017tf}.

A GBDT could be used for ranking by predicting a score for each agent. The final ranking is determined by these scores. A common approach to train the GBDT for ranking is to transform a list of objects to be ranked into a set of object pairs, and optimize a pairwise loss that penalizes the misorder of each pair with respect to the ground truth ranking; see~\cite{burges2010ranknet}.

In our system, we designed a convolutional neural network (CNN)~\cite{LeCun:1999:ORG:646469.691875} to improve the ranking quality. In the context of AVs, deep neural networks have been frequently used in recent years. We next review a subset:
\subsection{Deep Neural Networks for AVs}
CNNs and other variants have been used previously with AVs. For example, Bojarski {\it{et al}}~\cite{DBLP:journals/corr/BojarskiTDFFGJM16} trained a CNN to map raw pixels from a single front-facing camera directly to steering commands. Hecker {\it{et al}}~\cite{Hecker2018EndtoEndLO} used a deep network to map directly from the sensor inputs to future driving maneuvers.
Bansal {\it{et al}}~\cite{Ogale-RSS-19} trained a policy represented as a deep network for autonomous driving via imitation learning. They showed that their model can be robust enough to drive a real vehicle in some cases. They used a perception system that processes raw sensor
information to generate the network input: a top-down representation of the environment and
intended route, which is fed in sequentially to a recurrent neural network (RNN). We will adopt a similar top-down representation in our model. However, we will introduce a novel representation for time that will allow us to use a feed forward CNN, which is more efficient and easier to train. As our ranking system serves as a module in an AV software stack, we will use the output of a perception system, and design a CNN that takes the perception output as input.

\section{Data Generation}
To generate training data, we run the planner in simulation with computationally intensive models for all agents, and record which agents affected the final outcome. Some planners make this easier than others -- below, we first describe how we do this with a planner that explicitly attributes changes to a trajectory to specific agents; next, we provide a general method that is planner-agnostic. 

Our planner minimizes a cost function on plan $\eta$ under constraints $\mathcal{C}_i$ imposed by each Agent $i$; see for example~\cite{Fraichard583794}. We incrementally modify $\eta$ to satisfy all constraints, while keeping track of the active constraints~\cite{activeconstraints}. The magnitude by which $\mathcal{C}_i$ changes $\eta$ during optimization is recorded as an approximate indication of importance. In this paper, we categorize agents into 3 sets: {\it{Most relevant}}, {\it{Relevant}}, and {\it{Less relevant}} depending on the amount of change to $\eta$ using predetermined thresholds~\footnote{This approach could be applied with more fine-grained ranking, e.g. by looking at the exact deceleration/acceleration or the amount of geometric change. Combining speed and geometric reactions into a single ranking score is challenging. Thus, we used coarser buckets to avoid being biased to auto-labeling rules.}. 

 Although our experiments use the active constraints method from above, data generation can be made automatic for other kinds of planners as well -- even when treating a planner as a black box, we can still approximate the influence of different agents by planning multiple times. Denote by $f(\mathcal{A})$ the result of planning with computationally expensive models for the agents in the set $\mathcal{A}$. To approximate an agent's influence, we can look at the \emph{difference} in plans when we account for the agent versus when we do not. If $\eta=f(\emptyset)$ is the result of planning without any agents, and $\eta'=f(\{a_i\})$ is the result of planning with Agent $i$ only, then the value of the distance $d(\eta,\eta')$  is an indication of the agent's influence. Distances between trajectories can be measured with respect to the Euclidean inner product, with respect to difference in their costs for cost-based planners, or separately for geometry (using Dynamic Time Warping~\cite{Bemdt94usingdynamic}) and speed. Experimenting with different metrics and their ability to capture how well the planner does when not fully modeling a subset of agents is a topic for future work.

\section{Ranking Model}
We propose a hybrid approach which trains a GBDT on a combination of engineered and learned CNN features. The CNN uses a pairwise loss and consumes rich raw data encoding information about the dynamic scene in the form of top-down images. This enables the network to learn complex interactions between agents. Thus, as we will show in our experiments, the CNN helps improve the ranking quality. However, the CNN is designed to only cover $80 \times 80$ square meters around the AV to ensure its efficiency during onboard inference, as well as the high resolution of the input images. To rank all agents, we render the input images, run the CNN, and use its output to extract CNN features. Agents outside the CNN range take default CNN feature values. The GBDT takes the CNN features combined with engineered features as input to ensure that the agents outside the coverage range of the CNN can still be ranked using engineered features. In addition, the GBDT is efficient to allow re-ranking before every planning cycle. We next describe every component of our hybrid model.

\subsection{Convolutional Deep Network}

\begin{figure}[]
\center
\includegraphics[scale=.40,clip=true,angle=0]{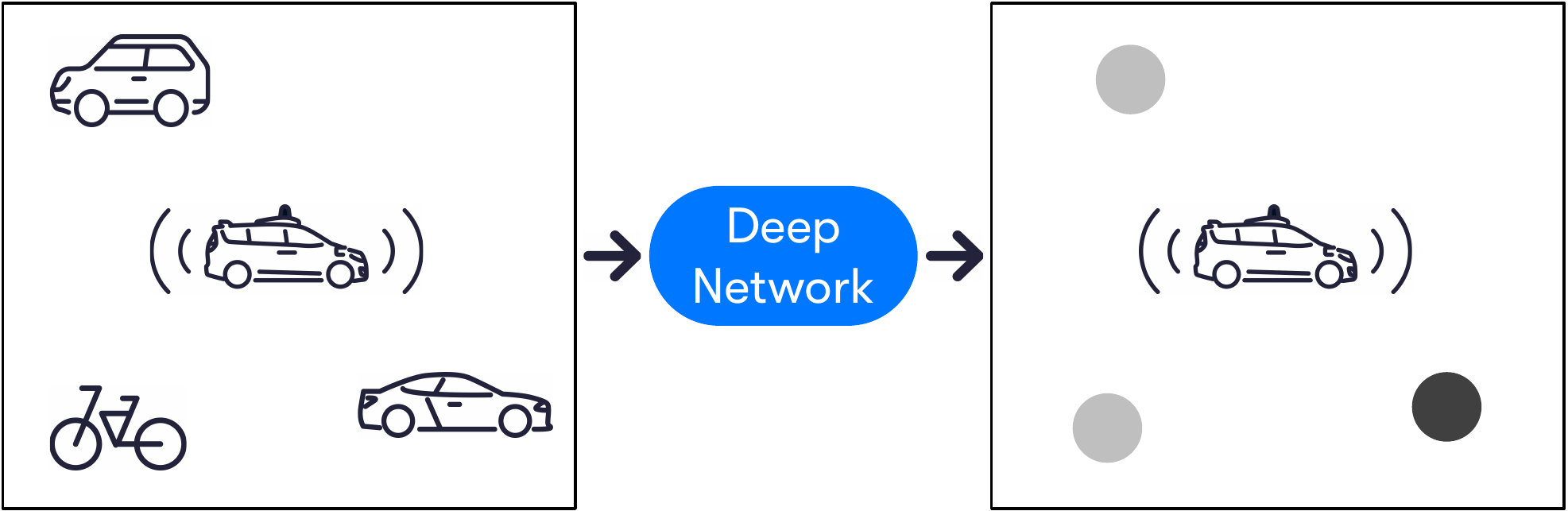}
\caption{A simplified block diagram showing the input and ground truth output of the deep network. The ground truth importance score of each agent is encoded in the output using the pixel intensity.
The colors of the output image were inverted in the figure for clarity.}
\label{fig:net_block}
\end{figure}

Our ranking CNN takes $200 \times 200$ pixels top-down images as input, where each pixel corresponds to an area of $40\times40$ square centimeters in the real world. All images are created with the AV being in the center location and heading up to have a common reference, ensuring corresponding pixels across images always match to the same location. This is key to fully realizing the benefits of our approach. The output of the network is an image with importance scores in the pixels where the agents are located as shown in Figure~\ref{fig:net_block}. All images use gray-scale values from $0$ to $255$ to encode different quantities. 

\begin{figure}[]
\center
\includegraphics[scale=.41,clip=true,angle=0]{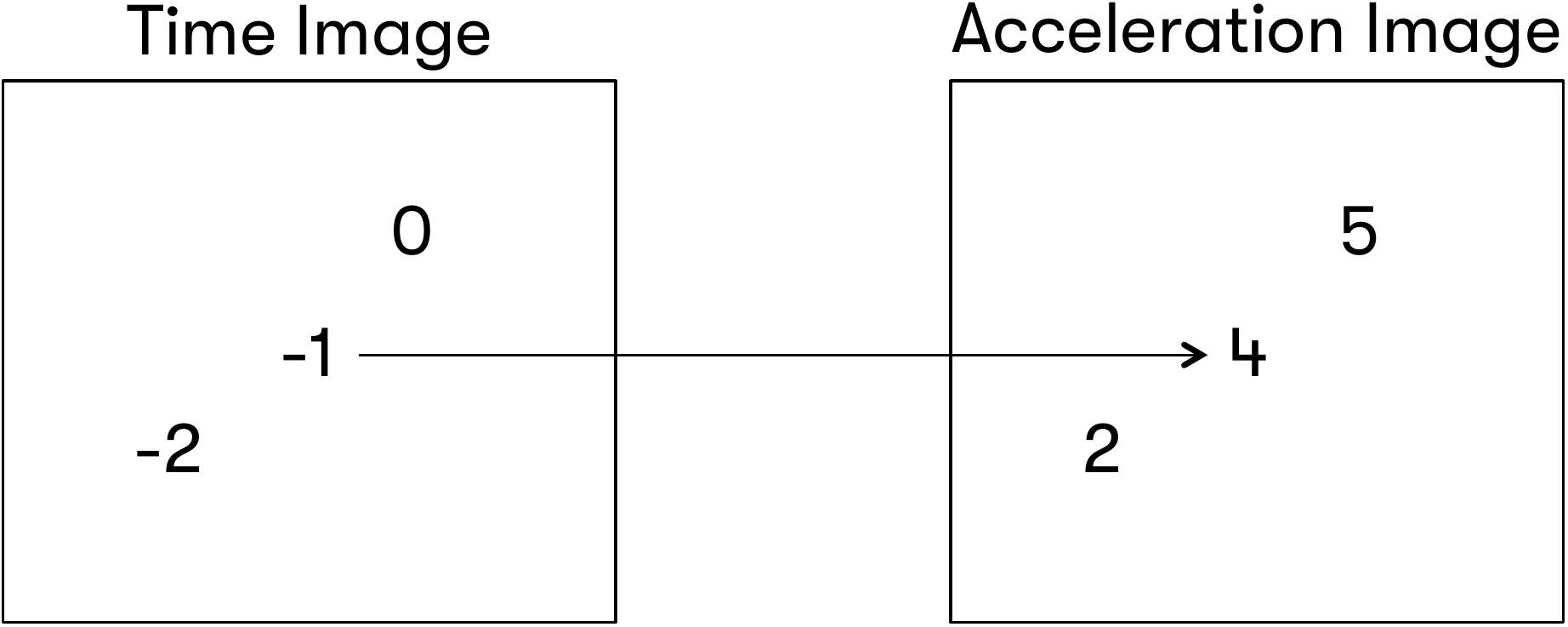}
\caption{The correspondence between the time channel and other input channels. Every pixel in the time image corresponds to the same pixel in the acceleration image. In this case, $2$ seconds ago (i.e. $-2$ in the time channel), the acceleration of the car at that location was $2$ as read from the acceleration channel; $1$ second ago, the acceleration was $4$, and now the acceleration is $5$.}
\label{fig:corr}
\end{figure}

To represent perceptual outputs over time compactly, we encode the time information in a separate channel. Let $t_1$ denote the time for which we want to compute the agent ranking. If the AV was at a particular location $(x, y)$ at time $t_2$, we plot the value $t_2 - t_1$ at the pixel corresponding to location $(x, y)$ in the time image. The time image can be used by the deep net to recover useful quantities over time, in combination with other input channels. For example, Figure~\ref{fig:corr} shows how the time channel can recover information about acceleration over time when combined with the acceleration channel. The semantics of the images is that every pixel in the time image corresponds to the same pixel in the acceleration image. Time in the past is represented with negative numbers, whereas future time is represented with positive numbers.

Using a separate time channel is useful for making the input more compact compared to a sequence of images which are equally spaced in time (i.e. video frames). To create the input images, we use the history of all agents and the AV, as well as the future predictions from the previous planning cycle, if available. Moreover, a quantity in an image encoded as a gray-scale pixel value is located at the pixel corresponding to the location where the quantity took place, or is predicted to take place. 

The input to the deep net is a set of $10$ top-down images as follows:

 \begin{enumerate}
\item The lanes in the scene in a separate image.
\item Three images for headings, velocities, and accelerations, respectively, of the AV at any point in time.
\item Time Image: every relative time of the AV in the pixel corresponding to where the AV was located, or is predicted to be located, at this relative time.
\item Three images for other agents' headings, velocities and accelerations, respectively, at any point in time. 
\item Other Agents Time Image: the relative times of the agents in the pixels corresponding to where the agents were located, or are predicted to be located, at these relative times.
\item Agents Image: the bounding boxes of all the agents located in $80 \times 80$ square meters around the AV at the current time.
 \end{enumerate}

\begin{figure}[]
\center
\includegraphics[scale=.4,clip=true,angle=0]{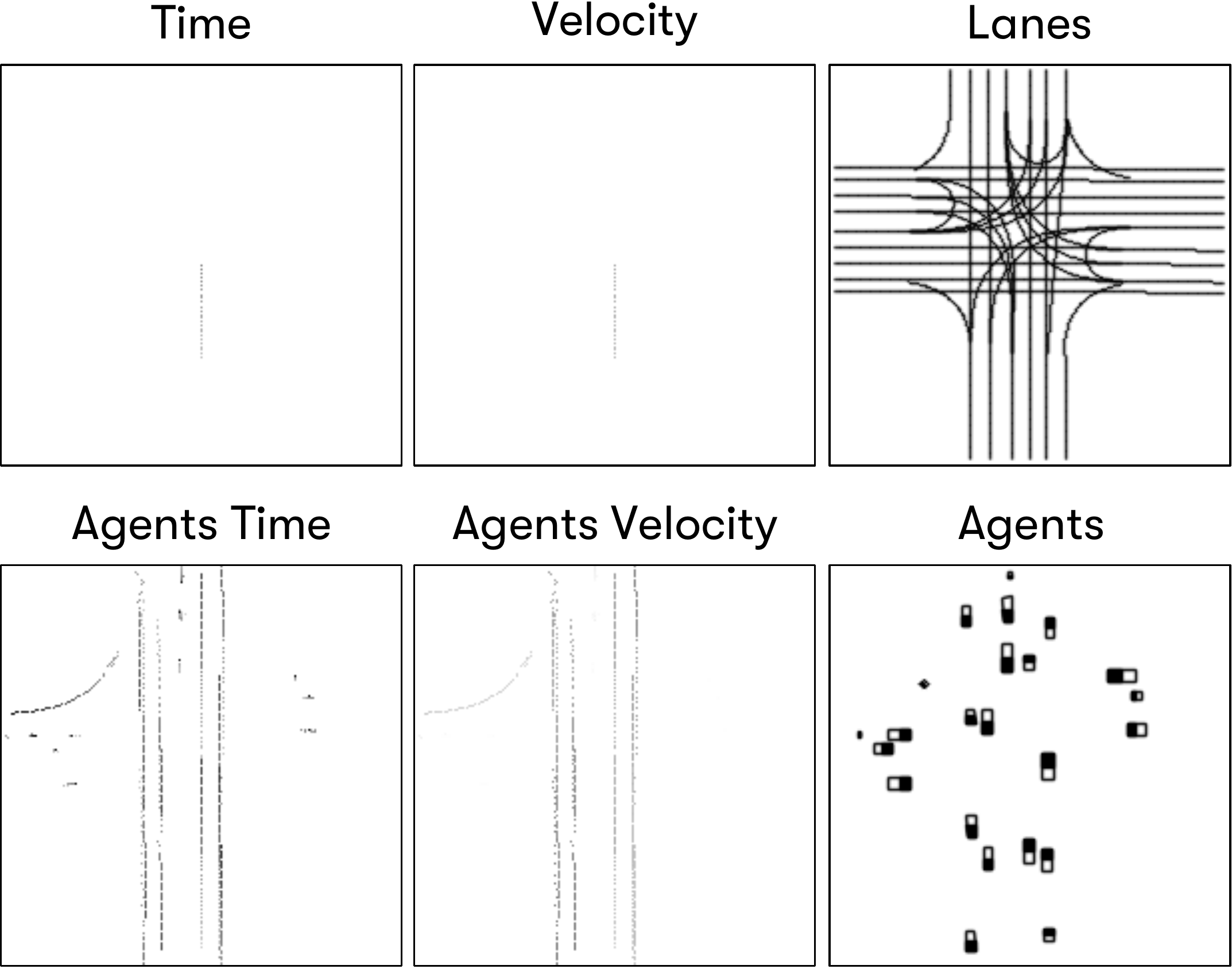}
\caption{A subset of the input channels provided to the deep network. We provide a separate time channel for each of the AV and the agents. The images were color inverted for clarity.}
\label{fig:images}
\end{figure}

Figure~\ref{fig:images} shows a subset of the images provided to the deep network.
The ground truth scores from the generated data are plotted in a label image, which is used by the loss function. Each score is written where the corresponding agent is located at the time ranking is computed~\footnote{In our implementation, we plotted a solid circle with the score instead of a single pixel to be able to inspect the images visually.}.

\begin{figure}[]
\center
\includegraphics[scale=.42,clip=true,angle=0]{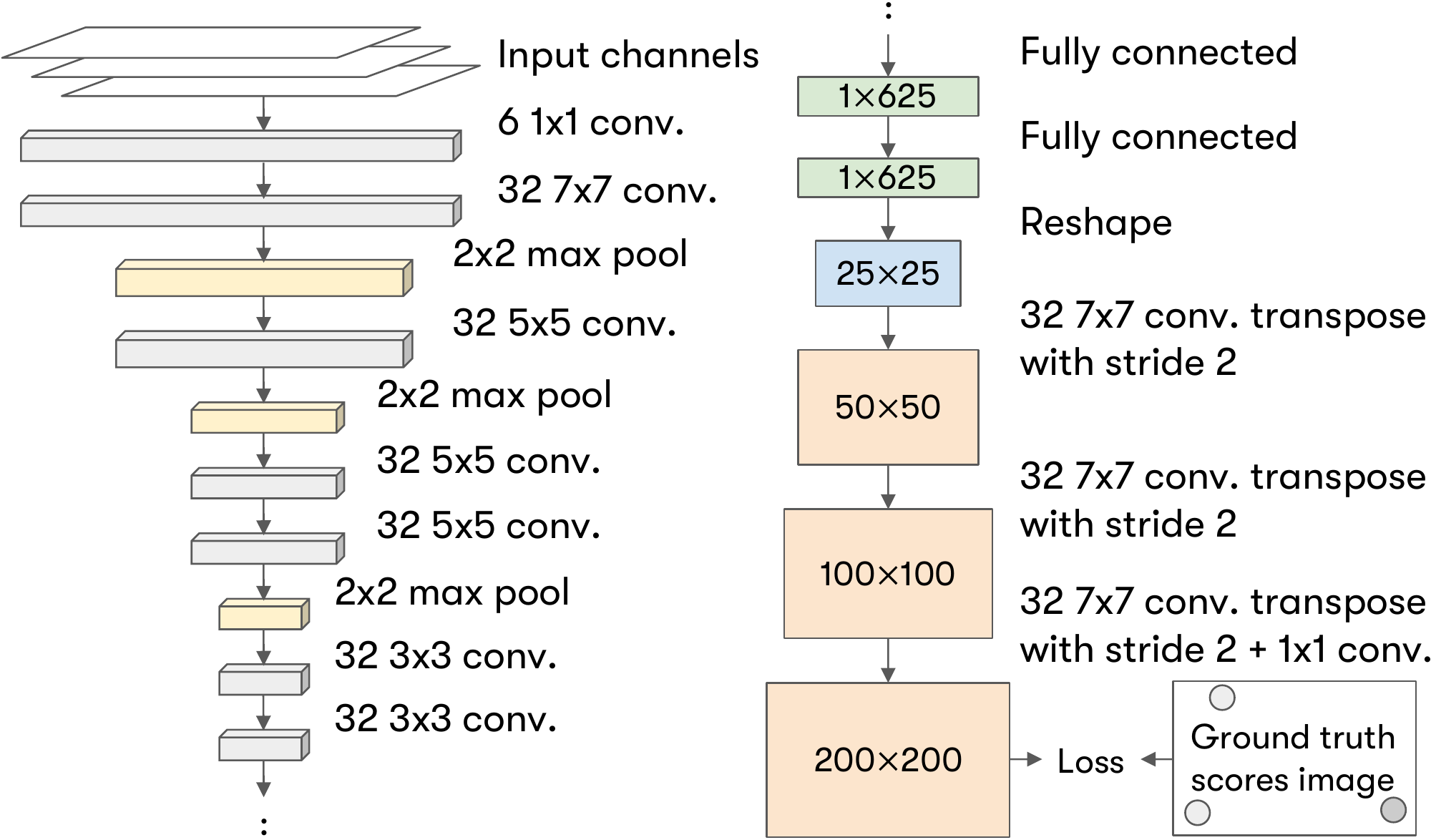}
\caption{The deep network architecture used to output importance scores for agents. The $2\times2$ max-pooling layer decreases the dimensions of its input tensor by half, which is depicted by the decrease in the tensor size in the figure. The convolution transpose operation doubles the dimensions of the input, allowing the recovery of the original $200 \times 200$ image size after applying fully connected layers to a smaller tensor (see {the tf.nn.conv2d\_transpose} documentation in TensorFlow). In some layers, we use multiple filters, for example in the first layer we use $6$ $1\times1$ filters, each of which is predicting its own output (not shown in the figure for clarity).}
\label{fig:arch}
\end{figure}

We use the network architecture in Figure~\ref{fig:arch} to predict a heat map with importance scores, and recover the agent scores from their locations.

\subsubsection{Loss Function}

\begin{figure}[]
\center
\includegraphics[scale=.65,clip=true,angle=0]{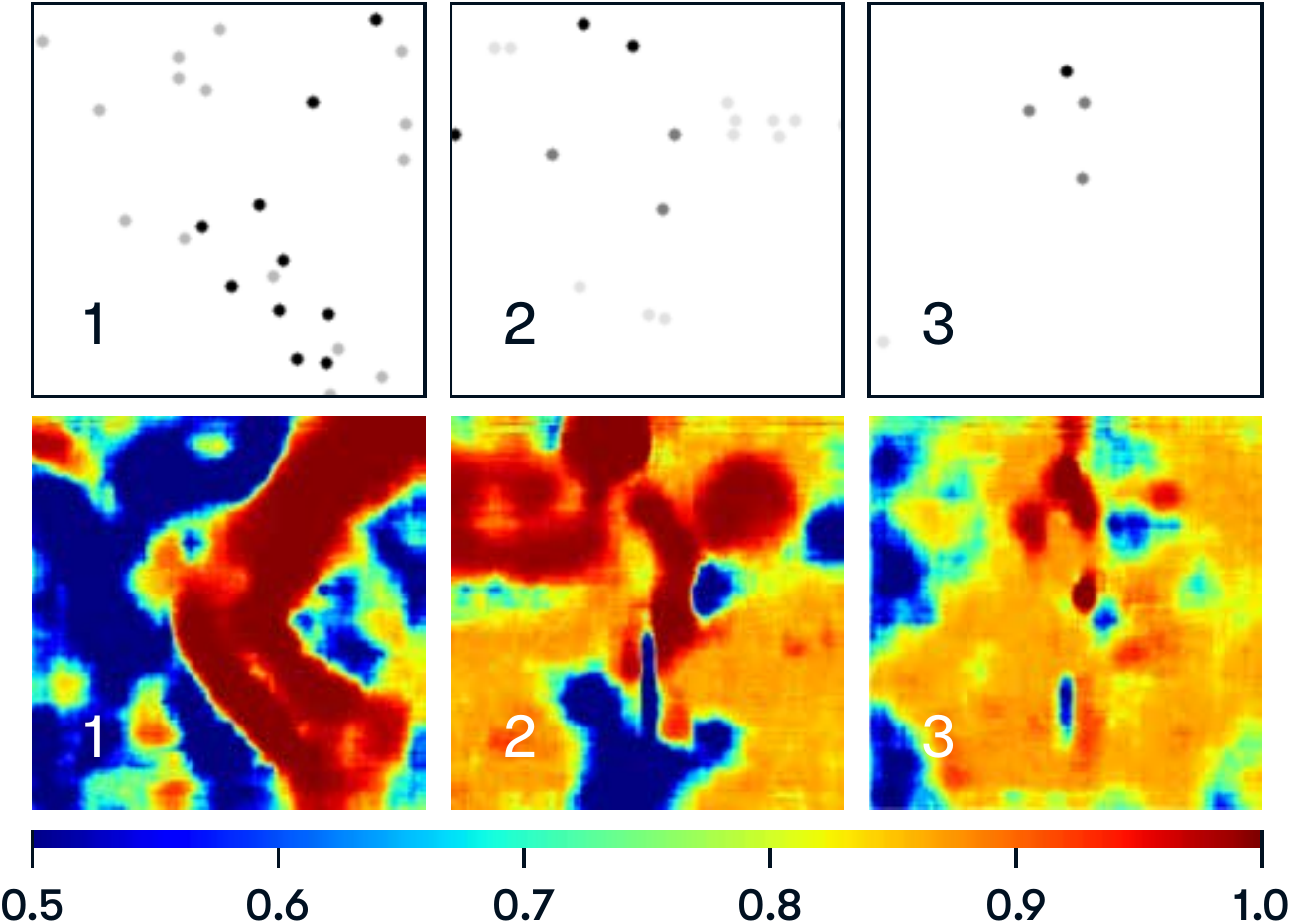}
\caption{Label images (top) and their corresponding predictions from the deep network (bottom). Darker labels depict more important agents. The pairwise loss only cares about relative scores. In the third image, for instance, the most important agent got a score of 0.996; the 3 neighboring agents got: 0.9959, 0.994, 0.987; and the bottom left agent got 0.51, producing a perfect ranking, despite being hard to rank visually via colors due to small differences.}
\label{fig:pairwise_ouput}
\end{figure}

Specific losses for ranking typically yield superior results over classification losses~\cite{burges2010ranknet}.
As such, we use a pairwise logistic loss applied to pairs of agents. Let the labels of Agents $i$ and $j$ be $l_i$ and $l_j$, and their scores from the deep net be $s_i$ and $s_j$, respectively. The loss incurred by a pair is $\log{(1 + e^{s_j-s_i})}$ if $l_i > l_j$, and 0 otherwise.
Figure~\ref{fig:pairwise_ouput} shows 3 label images and their corresponding predictions from the deep net.

\subsubsection{Deep Network Features}
The CNN score can be provided among the GBDT features. In this case, the agent's CNN feature is extracted from its corresponding location in the predicted image. In our implementation, we extracted multiple features for each agent from the neighborhood of its corresponding location. We have computed the maximum and average deep net scores in neighborhoods of $20 \times 20$, $10 \times 10$ and $5 \times 5$ around the agent's location. As the deep net predictions are not always perfect, extracting features from small neighborhoods can allow the model to be more robust. Agents outside the range take -1 default feature values.

\subsection {GBDT with Pairwise Loss}
The CNN features are provided to the GBDT as input after being augmented with the following hand-designed features:
\begin{enumerate}
\item The distance from the agent to the front of the AV.
\item Whether the agent is in front of the AV.
\item The current speed and acceleration of the agent.
\item Whether the agent is a cyclist, car, pedestrian, or child.
\item The distance from the agent to the AV's trajectory.
\item The time at which the AV's trajectory will be the closest to the agent.
\item The time needed by the agent to reach the AV's trajectory assuming a constant current acceleration.
\item The minimum time for the agent to collide with the AV given its current acceleration and the AV plan.
\end{enumerate}

The GBDT assigns a priority score to each agent represented by a feature vector. For training, one can naively utilize a pointwise classification loss. During inference, the list of agents is sorted based on the predicted scores. As we will show in the experiments, this model does not perform well, as the model is not optimized specifically for ranking.

To train the GBDT for ranking, we used the standard logistic loss over the difference of predictions for pairs of agents~\cite{burges2010ranknet}, sampled from the same planner iteration. We implemented several modifications to the vanilla algorithm, where an ensemble of decision trees are iteratively learned from an exhaustive list of object pairs~\cite{Burges05}:
\begin{enumerate}
\item \textit{Subsampling of pairs} 
In the classical form, for each list of ranked objects, an exhaustive list of pairs is used for training. Since global ranking does not exist in our problem (e.g. all objects in the {\it{Most relevant}} group are considered equally relevant, and all of them are more relevant than the objects in the {\it{Relevant group}}), we form the training data for GBDTs by sampling pairs of agents from different groups, and randomizing their order (i.e. $50\%$ chance the more important is the first). The label is positive if the first agent is more important than the other, and negative otherwise.

\item \textit{Layer-by-layer boosting}
To speed up learning, we used layer-by-layer boosting \cite{ponomareva2017compact}, which makes a boosting iteration for each layer of a tree, as opposed to a boosting iteration for the whole tree. This results in smaller ensembles reaching the same performance as ensembles of more weak learners \cite{ponomareva2017compact}, which is ideal for the efficient onboard inference we seek. 
\end{enumerate}

\section{EXPERIMENTAL RESULTS}

\begin{table*}[t]
\centering
\begin{tabular}{||c | c | c | c | c | c | c | c||} 
 \hline
 Approach & NDCG@1 & NDCG@3 & NDCG@5 & NDCG@10 & NDCG@20 & NDCG@30 & NDCG@40 \\ [0.5ex] 
 \hline\hline
 \textbf{Pairwise CNN + Pairwise GBDT} & \textbf{0.9291} & \textbf{0.9164} & \textbf{0.9171} & \textbf{0.9195} & \textbf{0.9201} & \textbf{0.9202} & \textbf{0.9183}  \\
 Pointwise CNN + Pairwise GBDT & 0.9217 & 0.9078 & 0.9088 & 0.9114 & 0.9110 & 0.9100 & 0.9077 \\
 Pairwise GBDT & 0.8792 & 0.8610 & 0.8651 & 0.8683 & 0.8639 & 0.8593 & 0.8546  \\
 Pairwise CNN + Pointwise GBDT & 0.8301 & 0.8530 & 0.8726 & 0.8840 & 0.8813 & 0.8790 & 0.8737  \\
 Pointwise CNN + Pointwise GBDT & 0.7862 & 0.8360 & 0.8581 & 0.8709 & 0.8667 & 0.8632 & 0.8588  \\
 Pointwise GBDT & 0.7444 & 0.8077 & 0.8273 & 0.8367 & 0.8307 & 0.8270 & 0.8230  \\
 Heuristics & 0.4743 & 0.5309 & 0.5582 & 0.5898 & 0.5993 & 0.6043 & 0.6055  \\ [1ex] 
 \hline
\end{tabular}
\caption{Average NDCG scores for different approaches. Each NDCG score is averaged across $\sim11$ million planning iterations from the test set.}
\label{table:ndcg}
\end{table*}

\begin{table*}[t]
\centering
\begin{tabular}{||c | c | c | c | c | c | c | c||} 
 \hline
 Approach & NDCG@1 & NDCG@3 & NDCG@5 & NDCG@10 & NDCG@20 & NDCG@30 & NDCG@40 \\ [0.5ex] 
 \hline\hline
 \textbf{Pairwise CNN + Pairwise GBDT} & \textbf{0.9748} & \textbf{0.9787} & \textbf{0.9782} & \textbf{0.9776} & \textbf{0.9770} & \textbf{0.9728} & \textbf{0.9679}  \\
 Pairwise CNN & 0.9723 & 0.9773 & 0.9768 & 0.9761 & 0.9754 & 0.9702 & 0.9647 \\
 Pointwise CNN + Pairwise GBDT & 0.9624 & 0.9658 & 0.9648 & 0.9635 & 0.9619 & 0.9564 & 0.9492 \\
 Pointwise CNN & 0.9408 & 0.9495 & 0.9483 & 0.9468 & 0.9433 & 0.9323 & 0.9150 \\
 Pairwise GBDT & 0.8968 & 0.8899 & 0.8862 & 0.8800 & 0.8816 & 0.8888 & 0.8957  \\ [1ex] 
 \hline
\end{tabular}
\caption{Average NDCG scores comparing different approaches trained and evaluated on agents in the $80\times80$ region. Each NDCG score is averaged across $\sim9$ million planning iterations from the test set where the CNN is applicable.}

\label{table:deepnet_analysis}
\end{table*}
We test the role of using learned models, utilizing CNN features, optimizing a pairwise loss, and adopting the hybrid approach. In particular, we manipulate some independent variables: 1) The type of loss used: pointwise vs pairwise. 2) The features used: from CNN, engineered, or hybrid. 3) For the hybrid approach: whether we are consistent in using the same type of loss. Hence, we evaluate the following approaches by learning models from the same training set and testing them on an unseen test set, both of which were generated by running the planner in simulation on real-world logged data:
  
\begin{enumerate}
  \item \textbf{Pointwise GBDT}: GBDT that uses the engineered features and utilizes a pointwise classification loss. 
  We use the same hand-designed features we have for our proposed model.
  \item \textbf{Pairwise GBDT}: GBDT that uses the engineered features while utilizing the pairwise loss.
  \item \textbf{Pointwise CNN}: CNN trained with the pointwise classification loss on agents within $80\times80$ square meters range.
  \item \textbf{Pairwise CNN}: CNN trained with the pairwise loss on agents within $80\times80$ square meters range.
  \item \textbf{Pointwise CNN + Pointwise GBDT}: GBDT with pointwise loss which uses the engineered features and features produced by the Pointwise CNN.
  \item \textbf{Pairwise CNN + Pointwise GBDT}: GBDT with pointwise loss which uses the engineered features and features produced by the Pairwise CNN.
  \item \textbf{Pointwise CNN + Pairwise GBDT}: GBDT with pairwise loss which uses the engineered features and features produced by the Pointwise CNN.
  \item \textbf{Pairwise CNN + Pairwise GBDT}: GBDT with pairwise loss which uses the engineered features and features produced by the Pairwise CNN.
  \item \textbf{Heuristics}: The heuristics based on the same engineered features, which mainly prioritize agents that are expected to overlap with the AV's trajectory sooner.
\end{enumerate}

The training set consists of planner reactions for agents from $\sim54$ million planning iterations whereas the test set consists of reactions from $\sim11$ million planning iterations extracted from different driving logs. The GBDTs in all approaches consist of $2$ trees, each of which has a maximum depth of $14$.

For a list of agents associated with the same planning iteration, we sort the list based on the scores produced by the model. To apply computationally intensive models to the top $K$ agents, we would like a metric that favors the appearance of highly relevant agents earlier in the sorted list. We use the NDCG~\cite{Jarvelin:2000:IEM:345508.345545} which penalizes the appearance of a highly relevant agent in lower positions as the graded relevance value is reduced logarithmically proportional to the position in the sorted list; see~\cite{Wang2013ATA}.

 To compute the NDCG, we first assign a numerical score $r(k)$ to every $k^{\text{th}}$ position of the sorted list. Namely, $r(k)$ is determined from the ground truth importance of the agent at the $k^{\text{th}}$ position: $2$ for most relevant agents, $1$ for relevant agents and $0$ for less relevant agents. The discounted cumulative gain (DCG) for the first $K$ agents in a sorted list is computed as:
\begin{equation}
 DCG@K = r(1) + \sum_{k=2}^{K}\frac{r(k)}{\log_{2}{k}}
\end{equation}
 This way DCG@K penalizes putting highly relevant agents at lower positions in a sorted list.

Across different planning iterations, the distribution of relevance can be very different. Therefore, we normalize DCG@K for each planning iteration. Namely, for each planning iteration, we compute an ideal DCG@K (i.e., IDCG@K) based on the list sorted by the ground truth importance scores. The normalized discounted cumulative gain for the top $K$ agents in a sorted list is computed as:

\begin{equation}
 NDCG@K=\frac{DCG@K}{IDCG@K}
\end{equation}

We compute the NDCG for each planning iteration for different approaches, and get the average for each $K$ across all planning iterations.

We test the following hypotheses:
$\mathcal{H}_1$: The learned models outperform the heuristics.
$\mathcal{H}_2$: The usage of the pairwise loss improves the ranking quality.
$\mathcal{H}_3$: The hybrid approach outperforms both the CNN and the GBDT alone.

Table~\ref{table:ndcg} shows the NDCG scores for different approaches. All machine learned models outperform the heuristics by $21\%$ to $45\%$ improvement ($\mathcal{H}_1$). It also demonstrates that both the deep net features and the pairwise loss improve the ranking quality. In terms of average NDCG@K, the performance of the GBDT with deep net features and pairwise loss is $8\%$ to $18\%$ better than the Pointwise GBDT with engineered features, across different K values, which agrees with $\mathcal{H}_2$ and $\mathcal{H}_3$.

In addition, in Table~\ref{table:2}, we evaluate how often an agent considered as one of the {\it{Most relevant}}, as defined in the data generation section, is placed at the first position in the sorted list determined by each approach. The hybrid approaches that use a Pairwise GBDT achieve the best results with $\sim 64\%$ improvement over the heuristics and $\sim 51\%$ improvement over the Pointwise GBDT. With deep
net features and pairwise loss, we are able to place one of the most
relevant agents at the first position in the ranked list $\sim 86\%$ of the time as opposed to only $\sim 36\%$ using the Pointwise GBDT, which agrees with our hypotheses.

\begin{table}[h!]
\centering
\begin{tabular}{||c | c||} 
 \hline
 Approach & Frequency \\ [0.5ex] 
 \hline\hline
 \textbf{Pairwise CNN + Pairwise GBDT} & \textbf{86.88\%}  \\
 Pointwise CNN + Pairwise GBDT & 86.10\% \\
 Pairwise GBDT & 77.97\%  \\
 Pairwise CNN + Pointwise GBDT & 55.62\%  \\
 Pointwise CNN + Pointwise GBDT & 43.44\%  \\
 Pointwise GBDT & 35.64\%  \\
 Heuristics & 21.89\%  \\ [1ex] 
 \hline
\end{tabular}
\caption{The frequency indicates how often an agent considered as one of the most relevant is placed at the first position in the sorted list.}
\label{table:2}
\end{table}

We also evaluate the ranking quality of the CNN alone compared to different GBDTs. The deep net by itself covers agents located in $80 \times 80$ square meters around the AV. Thus, for a fair comparison, the GBDTs are trained and tested on updated data sets where we remove agents on which the CNN is not applicable. All the GBDTs use the pairwise loss and the engineered features. Two GBDTs are provided with CNN features from the Pointwise and Pairwise CNNs, respectively.

 Table~\ref{table:deepnet_analysis} shows that using the deep net alone produces better ranking than the GBDT which is trained by the pairwise loss and only uses the engineered features. This is expected as the deep net is able to learn richer contextual information from the map and the raw input. Both the hybrid approach that has access to Pairwise CNN features, and the Pairwise CNN alone achieve the best results.

\begin{figure}[]
\center
\includegraphics[scale=.67,clip=true,angle=0]{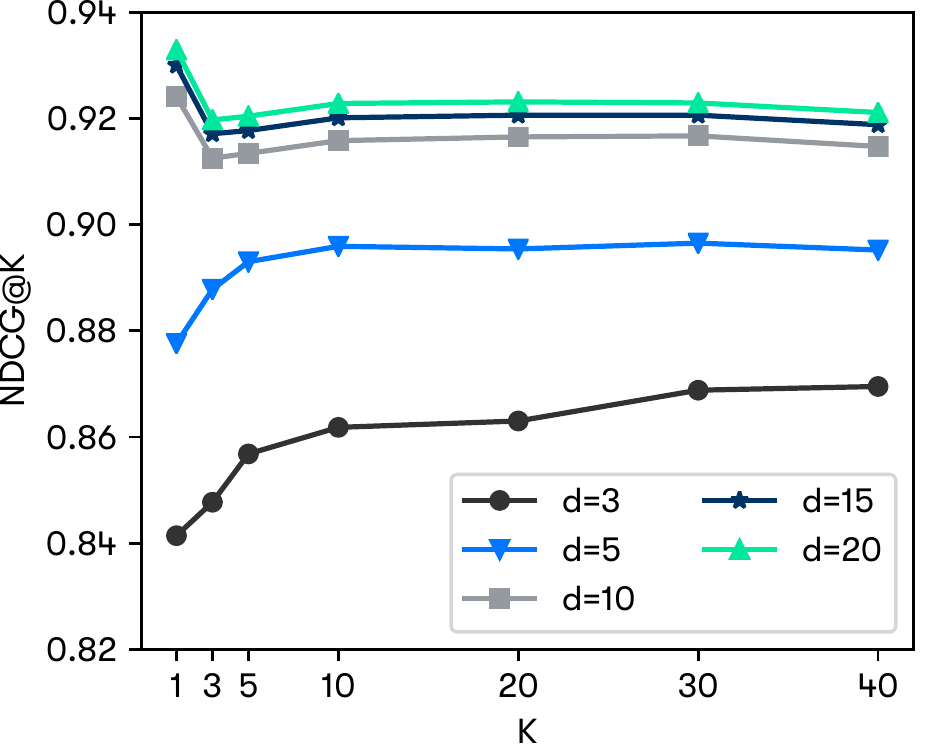}
\caption{Average NDCG scores for the hybrid approach with pairwise loss. The GBDT models consist of two trees with maximum depths from 3 to 20.}
\label{fig:model_complexity_depth}
\end{figure}

\begin{figure}[]
\center
\includegraphics[scale=.67,clip=true,angle=0]{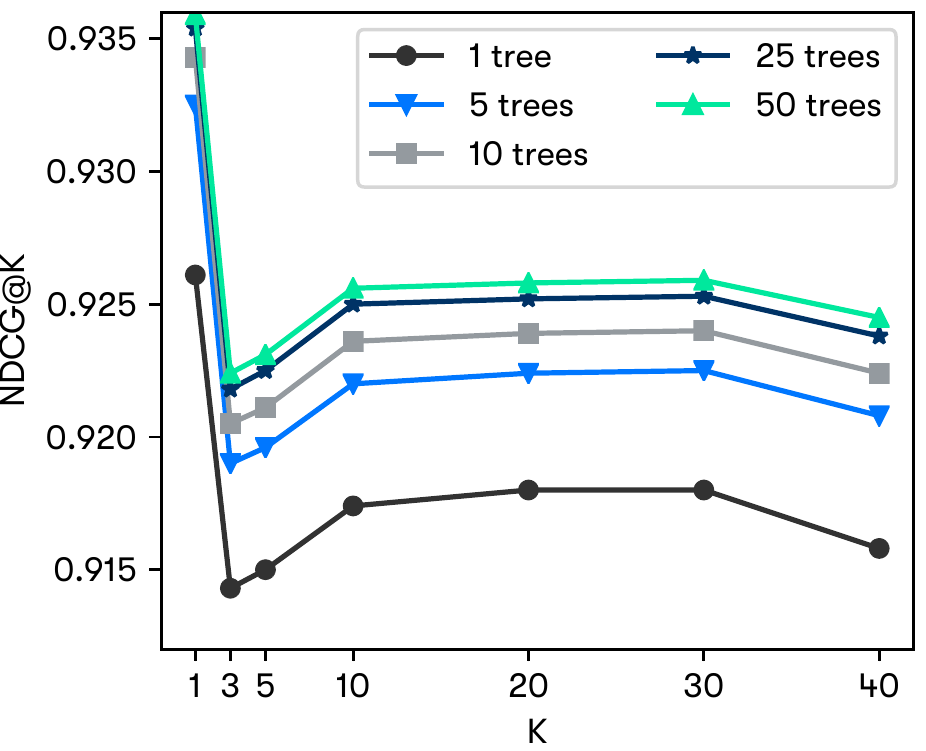}
\caption{Average NDCG scores for the hybrid approach with pairwise loss. The GBDT models consist of a number of trees varying from 1 to 50 with a fixed maximum depth of 14.}
\label{fig:model_complexity_num_trees}
\end{figure}

We study the correlation between the ranking quality and the complexity of the ranking model. We train GBDTs of different complexities, all of which use the pairwise loss and the hybrid features. We compare the average NDCG@K. As shown in Figures~\ref{fig:model_complexity_depth} and~\ref{fig:model_complexity_num_trees}, the ranking quality improves accordingly when we increase the maximum depth of each tree, or add more trees to the GBDT.

\begin{figure}[!t]

   \includegraphics[scale=.57,clip=true,angle=0]{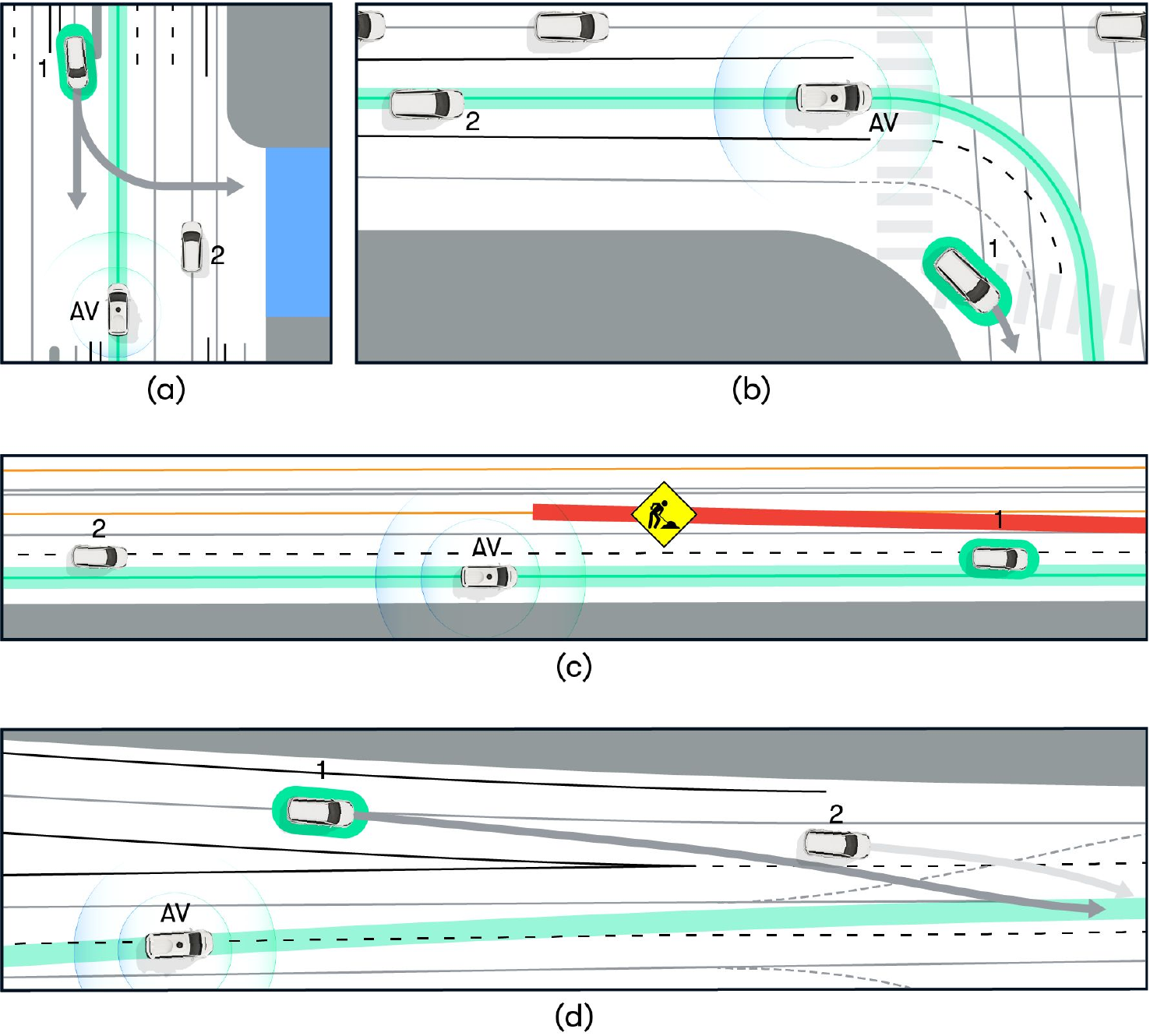}

  \caption{Examples of cases when our system ranks correctly. In all cases, Vehicle 1 is ranked as more important than Vehicle 2, and the speed limit is 45MPH. We describe each: (a) Vehicle 2 is traveling in parallel to the AV and, in reality, does not interact with it. Vehicle 1 is slowly driving forward at 12MPH and is about to turn into the driveway on its left. The decision of Vehicle 1 to yield to the AV or cross in front of it may cause a potential hazard. (b) The AV is planning for a right turn. The ranking model picks Vehicle 1 as the most relevant agent. Vehicle 1 is also making a right turn at 16MPH, and may make an immediate lane change after the turn. On the other hand, Vehicle 2 does not play an important role in the AV's decision making process. (c) Due to road construction, Vehicle 1, which is ahead of the AV by 40m, is traveling at 34MPH which constrains the AV's speed. (d) The AV is making a lane change to its left adjacent lane. Vehicle 2 is the closest agent to the AV's planned trajectory. However, in reality, Vehicle 1 is also making a lane change, and will cut in front of the AV before Vehicle 2.}

  \label{fig:qualitative_results} 
\end{figure}

Our system uses the hybrid approach with a GBDT consisting of only 2 trees and a maximum depth of 14, due to the strict requirement on the execution time of the AV system. 
Our model is able to rank thousands of agents within $\sim0.5$ milliseconds~\footnote{The deep net takes about $10$ milliseconds. To avoid blocking agent processing, we run it in parallel with the rest of the system, and use the deep net features from the previous cycle. This results in a negligible degradation in NDCG ($< 0.001$). If $160 \times 160$ images are used, the computational time grows to 18 milliseconds. In our experiments, rendering is done on an Intel Xeon W-2135 processor, and Nvidia Tesla V100 is used for inference.}.
Figure~\ref{fig:qualitative_results} shows interesting driving events where the learned ranking model is used.

\section{CONCLUSIONS}
\label{sec:conclusion}
We presented a ranking system that can be used onboard an AV to prioritize among agents in real time. Our system learns from automatically labeled data gathered by running the planner in simulation while logging causes of various reactions. We proposed a hybrid approach to ranking that utilizes engineered features, as well as features extracted from a convolutional neural network output, to optimize a pairwise loss. We showed the benefits of our system experimentally and analyzed various design choices. Qualitatively, our system appears to understand complex driving situations. The improvements in the ranking quality allow processing the most important agents with computationally intensive models and algorithms that can not be used with a large number of agents in real time. In our future work, we plan to evaluate various data generation mechanisms that are independent of a specific planner implementation.

\section*{Acknowledgments}
We thank Anca Dragan, Dragomir Anguelov, Nathaniel Fairfield, Jared Russell and Ryan Luna.





\bibliographystyle{IEEEtran}
\bibliography{IEEEabrv,references}
\end{document}